\documentclass[10pt,twocolumn,letterpaper]{article}

\usepackage[pagenumbers]{cvpr} 
\usepackage{multirow}
\usepackage{svg}
\definecolor{cvprblue}{rgb}{0.21,0.49,0.74}
\usepackage[pagebackref,breaklinks,colorlinks,allcolors=cvprblue]{hyperref}
\usepackage{float}

\def\paperID{*****} 
\def\confName{CVPR}
\def\confYear{2026}

\title{HQF-Net: A Hybrid Quantum-Classical Multi-Scale Fusion Network for Remote Sensing Image Segmentation}

\author{
Md Aminur Hossain$^{1, 2, \ast}$\quad
Ayush V. Patel$^{1}$  \quad
Siddhant Gole$^{2}$ \quad
Sanjay K. Singh$^{1}$ \quad
Biplab Banerjee$^{2}$\\
\\
$^{1}$Space Applications Centre, ISRO, India \quad
$^{2}$Indian Institute of Technology Bombay \\
\texttt{$\ast$Corresponding Author: md.aminurhossain@gmail.com}
}

\begin{document}
\maketitle
\begin{abstract}
Remote sensing semantic segmentation requires models that can jointly capture fine spatial details and high-level semantic context across complex scenes. While classical encoder--decoder architectures such as U-Net remain strong baselines, they often struggle to fully exploit global semantics and structured feature interactions. In this work, we propose \textbf{HQF-Net}, a hybrid quantum--classical multi-scale fusion network for remote sensing image segmentation. HQF-Net integrates multi-scale semantic guidance from a frozen DINOv3 ViT-L/16 backbone with a customized U-Net architecture through a Deformable Multiscale Cross-Attention Fusion (DMCAF) module. To enhance feature refinement, the framework further introduces quantum-enhanced skip connections (QSkip) and a Quantum bottleneck with Mixture-of-Experts (QMoE), which combines complementary local, global, and directional quantum circuits within an adaptive routing mechanism. Experiments on three remote sensing benchmarks show consistent improvements with the proposed design. HQF-Net achieves \textbf{0.8568} mIoU and \textbf{96.87\%} overall accuracy on LandCover.ai, \textbf{71.82\%} mIoU on OpenEarthMap, and \textbf{55.28\%} mIoU with \textbf{99.37\%} overall accuracy on SeasoNet. An architectural ablation study further confirms the contribution of each major component. These results show that structured hybrid quantum-classical feature processing is a promising direction for improving remote sensing semantic segmentation under near-term quantum constraints.
\end{abstract}
\section{Introduction}
\label{sec:intro}

Semantic segmentation assigns a semantic label to each pixel in an image, providing dense scene understanding for downstream decisions. Semantic segmentation is fundamental for remote sensing applications, including land cover mapping, urban growth monitoring, disaster assessment, precision agriculture and environmental monitoring \cite{zhu2017deep, guo2018semantic, martins2021semantic}. Compared to segmentation of natural images, remote sensing images typically have larger spatial extents, greater intra-class variance across regions and seasons, finer boundaries (e.g., roads, building edges), and a more pronounced multi-resolution structure in both texture and semantics.


In remote sensing segmentation, deep encoder-decoder architectures (including U-Net and similar architectures) provide strong baseline performance. However, because remote sensing images vary greatly across scales, textures, and acquisition methods, the performance of segmentation models is largely determined by the quality of the underlying feature representations. Some recent advancements in the utilization of self-supervised learning with vision transformers have produced strong visual representations that can create semantically informative and spatially consistent embedded representations using pre-trained features for use in dense prediction tasks. DINOv3 \cite{simeoni2025dinov3} is an example of these types of visual representation learning, where significant visual representations have been learned via self-distillation, resulting in feature maps containing meaningful high-level semantics while retaining their spatial structure. These representations have been shown to be beneficial.
 

Quantum machine learning (QML) is a computational paradigm that employs quantum phenomena, including superposition and entanglement, to create high-dimensional transformations in Hilbert spaces \cite{biamonte2017quantum, schuld2015introduction, du2025quantum}. The QML pipeline typically embeds classical input data into quantum states, and parameterized quantum circuits use unitary operators and measurements to learn the task-related representation of an object. This approach stems from the assumption that correlations may be more accurately captured in quantum feature spaces. However, current Noisy Intermediate-Scale Quantum (NISQ) devices provide limited qubit counts and restricted circuit depth, making direct quantum processing of high-dimensional imagery impractical. As a result, most vision-oriented QML approaches adopt hybrid quantum–classical designs, where small quantum circuits act as specialized modules within classical deep networks~\cite{hossain2026qmcnet}. Despite this progress, existing hybrid quantum vision models have largely focused on image-level tasks such as classification, while their application to dense prediction problems, particularly remote sensing semantic segmentation, remains limited.


We present \textbf{HQF-Net} (Hybrid Quantum-Classical Multi-Scale Fusion Network), a hybrid architecture for pixel-wise semantic segmentation of remote sensing imagery that combines robust transformer-based self-supervised representations with hybrid quantum-classical learning. HQF-Net utilizes a customized U-Net \cite{ronneberger2015u} style encoder-decoder, implementing quantum-enhanced feature interaction for better representation learning and enabling dense prediction in complex remote sensing scenes. In particular, the proposed framework is designed to address the research gap at the intersection of hybrid quantum learning, self-supervised visual representations, and dense remote sensing segmentation. The various components of HQF-Net provide a cohesive hybrid design for multi-scale representation fusion and quantum-assisted feature refinement in remote sensing segmentation. The main contributions of this work are:

\begin{enumerate}
    \item \textbf{Multi-scale fusion in a customized U-Net:} We develop a modified encoder-decoder U-Net to include multi-dimensional fusion modules by combining intermediate representations of features from a pre-trained DINOv3~\cite{simeoni2025dinov3} backbone (300M parameters), improving the model’s ability to capture both global and local semantic information
    
    \item \textbf{Quantum-augmented skip connections (QSkip):} We introduce quantum-enhanced skip connections, in which parameterized quantum circuits are designed to enhance complementary quantum feature interactions to enhance the capabilities of feature transfer for dense prediction.

    \item \textbf{Quantum bottleneck with QMoE:} We propose a bottleneck module that combines a quantum pre-transformation with a Quantum Mixture-of-Experts (QMoE), where a classical gating network adaptively mixes the outputs of multiple quantum expert circuits.

\end{enumerate}

By combining multi-scale DINOv3 features with quantum-augmented refinement, this work represents a step toward hybrid quantum-classical remote sensing semantic segmentation models.

\section{Related Work}
\label{sec:literature}



\textbf{Segmentation Models.} Ronneberger et al. \cite{ronneberger2015u} introduced U-Net as a segmentation architecture with symmetric encoding and decoding structures. In the U-Net, both the encoding and decoding regions are mirrored; i.e., their shapes and sizes match. The skip connections between the encoder and decoder enable the combination of deep semantic features with high-resolution spatial details. U-Net has achieved state-of-the-art accuracy on image-based datasets in medical imaging and has been widely applied to remote sensing tasks. D-LinkNet (Zhou et al. \cite{zhou2018d}), a modified version of U-Net that relies on a pretrained ResNet for encoding, achieved excellent results across various applications, further establishing U-Net as a benchmark architecture for image segmentation. Recently, attention mechanisms \cite{oktay2018attention} and transformer-based architectures have been incorporated into U-Net, enabling improved long-range dependency modeling and fine detail modeling, thereby improving performance on complex segmentation tasks.

\textbf{Quantum Machine Learning.} Over the past decade, quantum machine learning (QML) \cite{biamonte2017quantum} has emerged as a promising direction for developing quantum-enhanced learning models. Cong et al. \cite{cong2019quantum} introduced the Quantum Convolutional Neural Network (QCNN), a hierarchical quantum architecture composed of alternating convolution and pooling layers, and showed that it can learn expressive representations for complex classification tasks. Later, Henderson et al. \cite{henderson2020quanvolutional} proposed a hybrid variant in which small parameterized quantum circuits act as ``quanvolutional'' filters within a classical network. This line of work demonstrates that compact quantum circuits can be effectively integrated into classical architectures for feature extraction and representation learning.

Following these developments, QCNN-style circuits are particularly well suited to bottleneck architectures, where quantum modules are inserted after classical features have been highly compressed. This design is attractive under current hardware constraints because it reduces the data dimensionality before quantum processing. Prior work supports this strategy: Li et al. \cite{li2022image} combined a classical feature extractor with a quantum classifier on EuroSAT and achieved performance comparable to a classical model. Similarly, ~\cite{hossain2026qmcnet} demonstrated the efficacy of data-aware quantum representations for remote sensing classification within hybrid frameworks. Since deep quantum models also face optimization difficulties due to the barren plateau problem \cite{pesah2021absence}, recent segmentation studies have placed quantum circuits at the bottleneck of U-Net-like architectures to refine compact features before decoding \cite{jain2025qufex, wang2025qc, halab2025qu}. This motivates the hybrid quantum bottleneck design used in HQF-Net.

\begin{figure*}[t!]
    \centering
    \includegraphics[width=2.0\columnwidth]{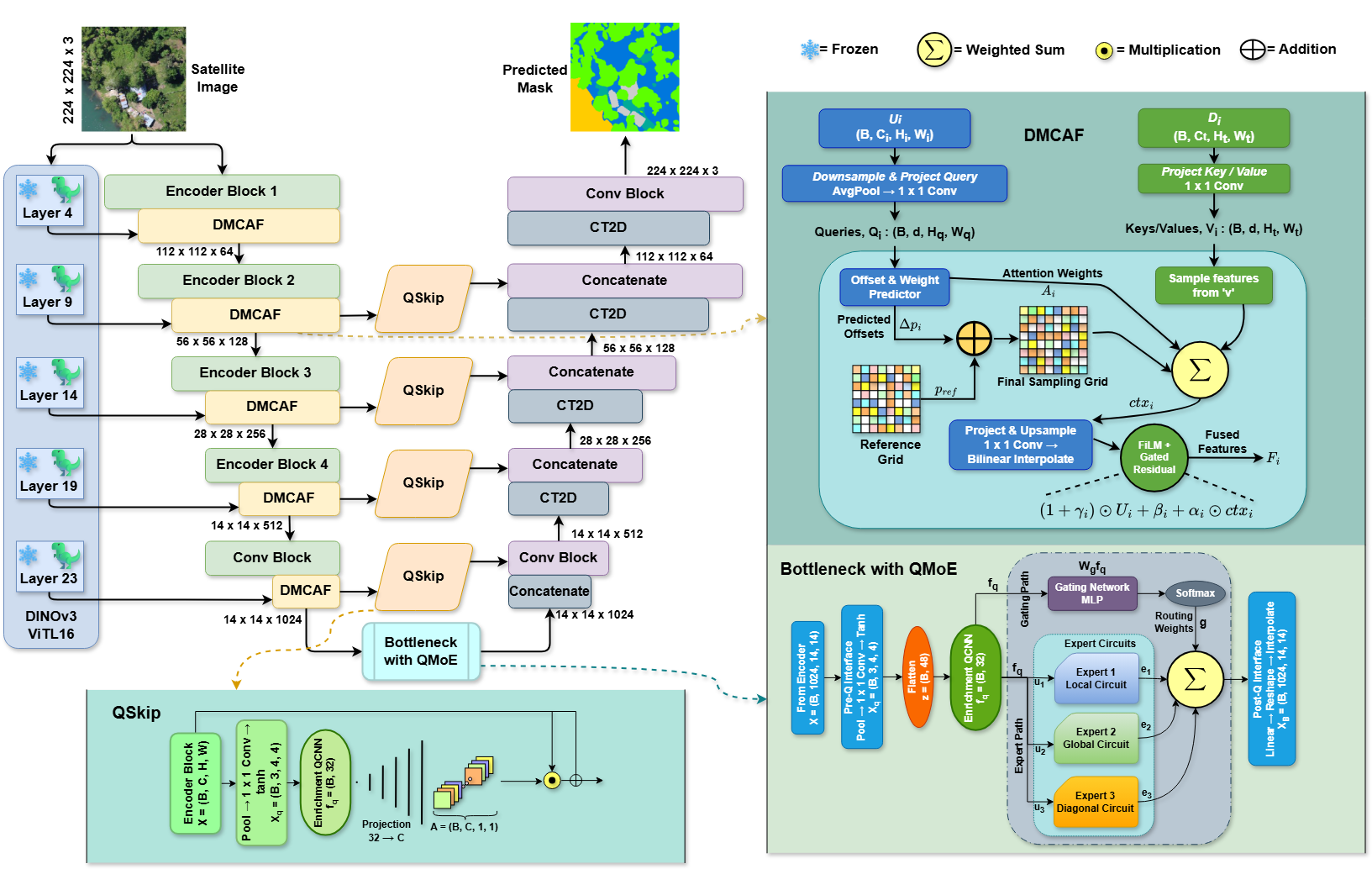}
    \caption{Overview of the proposed HQF-Net architecture. The model integrates a frozen DINOv3 ViT-L/16 encoder with Deformable Multi-Scale Cross-Attention Fusion (DMCAF) module. The main hybrid components are Quantum Skip (QSkip) refinement blocks and a bottleneck with Quantum Mixture-of-Experts (QMoE) block that adaptively combines local, global, and diagonal quantum experts. CT2D denotes transposed convolution.
}
    \label{fig:architecture}
\end{figure*}


Mixture-of-Experts (MoE) models improve task performance by flexibly allocating computation across multiple specialized experts \cite{riquelme2021scaling}. The current state of the art further improves efficiency through sparse activation, selecting only a small subset of experts for each input, thereby reducing computational cost while maintaining strong performance on large-scale tasks such as image segmentation \cite{lepikhin2020gshard}. In segmentation, this selective specialization is particularly valuable because different experts can focus on distinct visual patterns or semantic structures, making sparse MoE well-suited for complex domains such as remote sensing (RS) and medical imaging (MI) \cite{riquelme2021scaling}. This specialization-based view motivates our use of an MoE design within HQF-Net, where different quantum experts are intended to capture complementary feature dependencies.

\section{Proposed Methodology}
\label{sec:methodology}

HQF-Net (Fig.~\ref{fig:architecture}) is a hybrid architecture that combines quantum circuits with a U-Net based semantic segmentation network. Quantum circuits enhance feature maps present within the Skip Connection and Bottleneck layers, while also implementing a Mixture of Experts (MoE) approach \cite{dai2024deepseekmoe} at the Bottleneck layer for improved segmentation. A pre-trained DINOv3 Vision Transformer guides the classical encoder to learn the feature space at multiple scales. We used these pretrained encoder because of its ability to learn representations through self-supervised methods and trained on satellite images with 300 million parameters only. It is able to generate a significantly greater number of more robust and more transferable features compared to traditional backbones. This section provides detailed information on HQF-Net's architecture and its components.

HQF-Net preserves the macro-level encoder–decoder topology of the classical U-Net while replacing key internal components with quantum-enhanced modules. The framework consists of five major stages: a classical encoder, deformable multiscale cross-attention fusion, quantum-enhanced skip refinement, a quantum Mixture-of-Experts bottleneck,  and a classical decoder.

\subsection{Classical Encoder Blocks}
These blocks generate a set of feature maps using depthwise separable convolutions in a parameter-efficient manner, from the input image patch at progressively lower spatial resolutions. This encoding process produces compact feature representations for subsequent quantum processing.

\subsection{Deformable Multiscale Cross-Attention Fusion}
To align high-level semantic representations from DINOv3 with dense spatial features from the U-Net encoder, we introduce a lightweight adapter termed Deformable Multiscale Cross-Attention Fusion (DMCAF). Instead of naively upsampling DINOv3 features and concatenating, DMCAF performs \emph{query-anchored deformable cross-attention} followed by FiLM-gated residual injection. This design is inspired by deformable attention mechanisms \cite{zhu2020deformable}.
    
Let the encoder feature of U-Net at stage $i$ be $U_i \in \mathbb{R}^{B \times C_i \times H_i \times W_i}$ and the corresponding pretrained DINOv3 ViT's feature $D_i \in \mathbb{R}^{B \times C_t \times H_t \times W_t},$ where $B$ denotes the batch size. Instead of directly upsampling $D_i$ and fusing it with $U_i$, we adopt an \textit{align-then-inject} strategy using deformable cross-attention.
The encoder feature is first spatially reduced using stride $s_i$:
\begin{equation}
    U_i^q = \mathrm{AvgPool}_{s_i}(U_i),
\end{equation}
where
\begin{equation}
    H_q = \frac{H_i}{s_i}, \quad W_q = \frac{W_i}{s_i}.
\end{equation}
The reduced feature is projected into a shared attention space:
\begin{equation}
    Q_i = \mathrm{Conv}_{1\times1}(U_i^q),
\end{equation}
\begin{equation}
    Q_i \in \mathbb{R}^{B \times d \times H_q \times W_q}.
\end{equation}
Transformer features are similarly projected:
\begin{equation}
    V_i = \mathrm{Conv}_{1\times1}(D_i),
\end{equation}
\begin{equation}
    V_i \in \mathbb{R}^{B \times d \times H_t \times W_t},
\end{equation}
where $d$ is the shared attention dimension. For each query location $x$, a reference location $p^{ref}(x)$ is defined by mapping the query grid to the transformer grid. An offset predictor produces $\Delta p_i \in \mathbb{R}^{B \times K \times H_q \times W_q \times 2},$ and attention weights $A_i \in \mathbb{R}^{B \times H \times H_q \times W_q \times K},$ where $K$ is the number of sampling points. The sampling locations are:
\begin{equation}
    p_i^{(k)}(x) = p^{ref}(x) + \Delta p_i^{(k)}(x).
\end{equation}
Context aggregation is performed via bilinear sampling:
\begin{equation}
    \mathrm{ctx}_i(x)
    =
    \mathrm{Concat}_{h=1}^{H}
    \left(
    \sum_{k=1}^{K}
    A_{i,h}^{(k)}(x)\,
    V_{i,h}\big(p_{i,h}^{(k)}(x)\big)
    \right)
\end{equation}
Thus,
\begin{equation}
    \mathrm{ctx}_i \in \mathbb{R}^{B \times d \times H_q \times W_q}.
\end{equation}
This reduces complexity from dense cross-attention $\mathcal{O}(H_q W_q H_t W_t)$ to sparse attention $\mathcal{O}(H_q W_q K)$.
The contextual feature is projected back to the encoder channel space:
\begin{equation}
    \tilde{\mathrm{ctx}}_i  =\mathrm{Conv}_{1\times1}(\mathrm{ctx}_i),
\end{equation}
\begin{equation}
    \tilde{\mathrm{ctx}}_i\in\mathbb{R}^{B \times C_i \times H_q \times W_q}.
\end{equation}
It is then upsampled to match the encoder resolution:
\begin{equation}
        \hat{\mathrm{ctx}}_i=\mathrm{Bilinear}(\tilde{\mathrm{ctx}}_i),
\end{equation}
\begin{equation}
    \hat{\mathrm{ctx}}_i\in\mathbb{R}^{B \times C_i \times H_i \times W_i}.
\end{equation}
Modulation parameters are computed via global pooling:
\begin{equation}
    \gamma_i, \beta_i, \alpha_i =\Psi\big(\mathrm{GAP}(\hat{\mathrm{ctx}}_i)\big),
\end{equation}
where $\gamma_i, \beta_i, \alpha_i \in \mathbb{R}^{B \times C_i}$.
The final fused feature is:
\begin{equation}
        F_i
        =
        (1 + \gamma_i)\odot U_i
        +
        \beta_i
        +
        \alpha_i \odot \hat{\mathrm{ctx}}_i,
\end{equation}
\begin{equation}
        F_i
        \in
        \mathbb{R}^{B \times C_i \times H_i \times W_i},
\end{equation}
where $\odot$ denotes channel-wise modulation.

\subsection{Bottleneck with Mixture-of-Experts (QMoE)} 
\label{sec:qmoe}
To enhance the representational power at the bottleneck stage, we introduce a Bottleneck with QMoE module that integrates multi-scale quantum feature enrichment with a quantum Mixture-of-Experts (QMoE) routing mechanism.
The output of the final encoder stage is:
\[
        \mathbf{X} \in \mathbb{R}^{B \times 1024 \times 14 \times 14}.
\]
Since quantum circuits require fixed-size inputs, we first compress the encoder feature map spatially and channel-wise:
\begin{equation}
        \mathbf{X}_{q} =\tanh\left(\text{Conv}_{1\times1}
        \big(\text{AdaptiveAvgPool}_{S \times S}(\mathbf{X})\big)\right),
        \label{eq:compression}
\end{equation}
where,
\begin{equation}
        \mathbf{X}_{q} \in \mathbb{R}^{B \times 3 \times 4 \times 4}
        \label{eq:downsampled_feature}
\end{equation}
The tensor is flattened:
\begin{equation}
        \mathbf{z} \in \mathbb{R}^{B \times 48}
        \label{eq:flatten}
\end{equation}
Each sample vector \( \mathbf{z}_b \in \mathbb{R}^{48} \) is reshaped into
\begin{equation}
        \mathbf{z}_b \rightarrow \mathbb{R}^{16 \times 3},
        \label{eq:reshaped}
\end{equation}
where each of the \(N=16\) qubits receives three feature components. The encoding prepares the quantum state:
\begin{equation}
        |\psi\rangle =\prod_{i=1}^{16}
        R_X(\pi z_{i,1})
        R_Y(\pi z_{i,2})
        R_Z(\pi z_{i,3})
        |0\rangle.
        \label{eq:encoding}
\end{equation}
A dedicated enrichment multiscale quantum circuit applies: Horizontal grid convolutions, Vertical grid convolutions, Global cyclic entanglement.
The enriched quantum features are extracted via expectation measurements:
\begin{equation}
        \mathbf{f}_q =
        \left[
        \langle Z_i \rangle,
        \langle X_i \rangle
        \right]_{i=1}^{16}
        \in \mathbb{R}^{32}
        \label{eq:measurement}
\end{equation}
A classical gating network dynamically routes information:
\begin{equation}
        \mathbf{g} =
        \text{Softmax}(W_g \mathbf{f}_q),
        \quad
        \mathbf{g} \in \mathbb{R}^{3}
\end{equation}
Each expert receives a learned projection:
\begin{equation}
        \mathbf{u}_k = W_k \mathbf{f}_q, \quad where \quad k \in \{1,2,3\},
\end{equation}
followed by a specialized quantum circuit:
\begin{equation}
        \mathbf{e}_k = \mathcal{Q}_k(\mathbf{u}_k)  \quad where \quad  k \in \{1,2,3\},
        \quad
        \mathbf{e}_k \in \mathbb{R}^{32}
\end{equation}
The three experts are designed to capture complementary correlations: Local spatial correlations, global entanglement patterns, structured diagonal dependencies.

The final quantum bottleneck representation is obtained via weighted aggregation:
\begin{equation}
        \mathbf{f}_{\text{MoE}} =
        \sum_{k=1}^{3}
        g_k \mathbf{e}_k.
\end{equation}
The mixed quantum representation is projected to the desired bottleneck dimension:
\begin{equation}
        \mathbf{b} = W_o \mathbf{f}_{\text{MoE}},
        \quad
        \mathbf{b} \in \mathbb{R}^{C_{\text{out}}}.
\end{equation}
Finally, the vector is reshaped and broadcast back to spatial resolution:
\begin{equation}
        \mathbf{X_B} \in
        \mathbb{R}^{B \times C_{\text{out}} \times 14 \times 14}.
\end{equation}

\subsection{Quantum-Enhanced Skip Connections}
To enhance feature refinement within skip connections, we introduce a Quantum Skip Attention (QSkip) module that adaptively modulates encoder features before fusion with the decoder. Let the encoder feature map be:
\begin{equation}
        \mathbf{X} \in \mathbb{R}^{B \times C \times H \times W}.
\end{equation}
To interface with the quantum circuit, spatial and channel compression is first applied as described in Section~\ref{sec:qmoe} module (Eqs.~\ref{eq:compression}-\ref{eq:measurement}) to get $f_q$.
These 32 quantum descriptors from $f_q$ are projected back to channel dimension:
\begin{equation}
        \mathbf{a} =
        \sigma\left(
        W_a \mathbf{f}_q
        \right),
        \quad
        \mathbf{a} \in \mathbb{R}^{C},
\end{equation}
where \( \sigma(\cdot) \) is the sigmoid activation.
The attention vector is reshaped and broadcast spatially:
\begin{equation}
        \mathbf{A} \in \mathbb{R}^{B \times C \times 1 \times 1}.
\end{equation}
Feature refinement is then performed via multiplicative modulation with residual preservation:
\begin{equation}
    \mathbf{Y} = \mathbf{X} \odot \mathbf{A} + \mathbf{X}.
\end{equation}
Unlike classical squeeze-and-excitation blocks, QSkip uses quantum feature transformations to extract global inter-channel correlations from compressed spatial descriptors. The residual formulation ensures stable gradient flow while enabling quantum-guided channel recalibration.

\subsection{Classical Decoder Blocks}
The decoder's output, the final segmentation mask, is produced from the Quantum-enhanced skip connections and the processed bottleneck feature map, using transposed convolutions (upsampling) and convolutional blocks to reconstruct the input’s spatial dimensions while preserving high-resolution details. Each skip feature is fused with its corresponding decoder stage.

\begin{figure*}[h!]
    \centering
    \includegraphics[width=\linewidth]{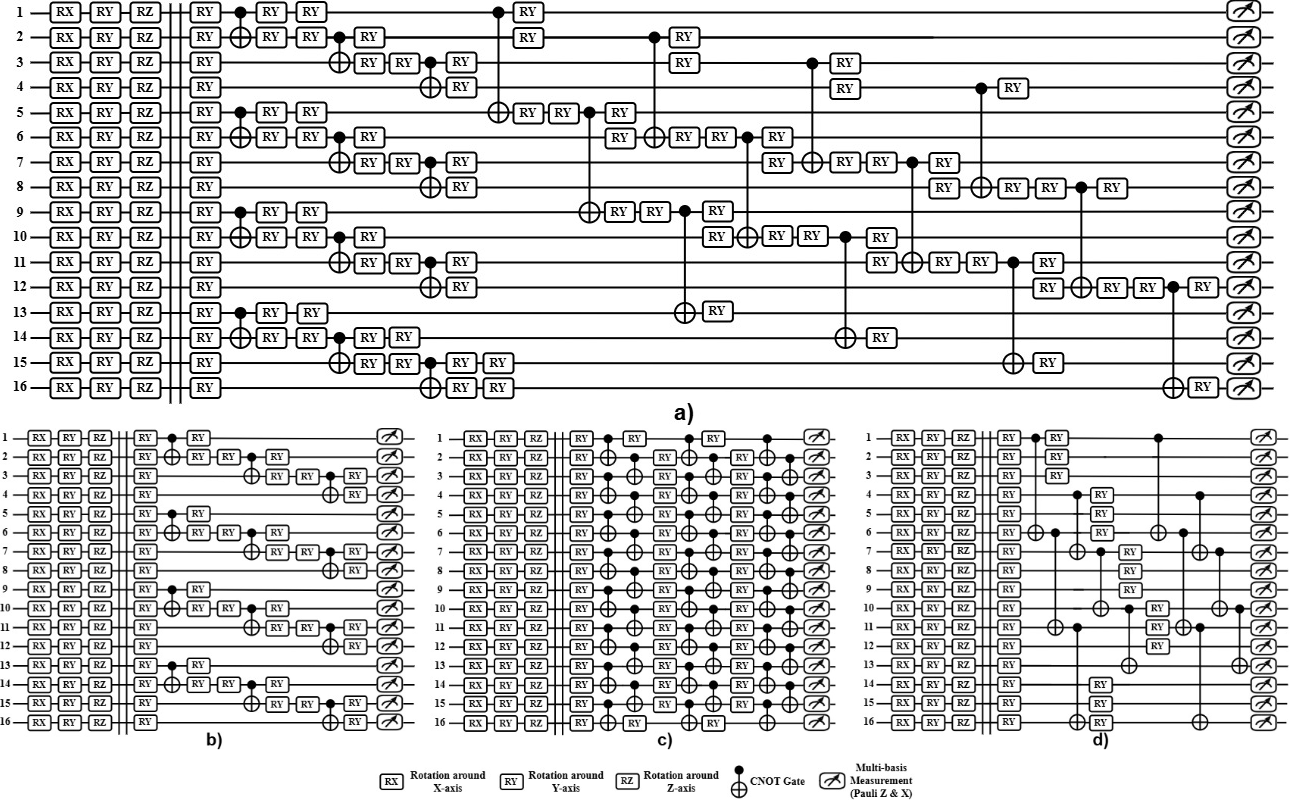}
    \caption{Overview of Quantum Circuits used in HQF-Net - a) Enrichment Multiscale Circuit, b) Local Circuit, c) Global Circuit, and d) Diagonal Circuit}
    \label{fig:allcirc}
\end{figure*}

\subsection{Quantum Feature Processing Modules}
The quantum modules are integrated at two key locations in the architecture: the QMoE bottleneck and the QSkip. Here, we summarize the main circuit designs, while full architectural and implementation details are provided in the supplementary material.

\textbf{1. The Enrichment Multi-Scale Circuit:}
This circuit (Fig.~\ref{fig:allcirc}-a) is designed to capture both local and global feature correlations in the input representation. It first applies localized operations on neighboring qubits to model fine-grained spatial patterns, followed by a global mixing stage that entangles all qubits to propagate information across the full feature representation. In HQF-Net, this circuit is used within the QSkip module and before the QMoE block.

\textbf{2. Expert Circuits:} The QMoE bottleneck employs three complementary expert circuits to capture local, global, and directional dependencies. The \emph{Localist} circuit (Fig.~\ref{fig:allcirc}-b) models fine-grained local structure through entanglement between neighboring qubits, making it suitable for short-range patterns such as edges and textures. The \emph{Globalist} circuit (Fig.~\ref{fig:allcirc}-c) uses broader entanglement and rotation operations to capture non-local interactions and global contextual structure. The \emph{Diagonal} circuit (Fig.~\ref{fig:allcirc}-d) applies parameterized rotations and CNOT gates over diagonally related qubits to model directional and structured feature relationships beyond purely local or global connectivity.
\section{Experimental Results}
\label{sec:results}

\subsection{Datasets and Experimental Setup}
This section describes the datasets, and training configurations used to evaluate the proposed architecture. The datasets used to evaluate the models were sourced from three complex remote sensing datasets and served as the basis for evaluating model performance across different segmentation tasks.

\subsubsection{Datasets}
For evaluation, we use three high-resolution datasets of aerial and satellite imagery. \textbf{LandCover.ai}~\cite{boguszewski2021landcover} contains aerial imagery of Poland at 25–50 cm/pixel, focusing on land-use mapping with five classes: Building, Forest, Water, Road, and Background. The large orthophotos are uniformly processed into $512 \times 512$ image-mask tiles. \textbf{OpenEarthMap}~\cite{xia2023openearthmap} is a global collection of high-resolution satellite imagery from multiple regions, each with its own training, validation, and testing splits; original images are $1000 \times 1000$ pixels. \textbf{SeasoNet}~\cite{kossmann2022seasonet} is a multi-temporal, multi-source UAV dataset over agricultural land; for our experiments, we use a subset of winter and summer images. SeasoNet contains 33 semantic classes, with images conveniently pre-assembled into $120 \times 120$ image-mask pairs.


\subsubsection{Experimental Setup}
The models were trained for 100 epochs using Adam~\cite{kingma2014adam} optimizer with a learning rate of 0.0001 and a standard CrossEntropyLoss function. The input size was $224\times224\times3$. A batch size of 32 was used for all experiments, determined by the available GPU memory. The models were trained on NVIDIA A100 80GB PCIe GPUs.

\subsection{Results and Discussion}
This section presents both quantitative and qualitative evaluations of our proposed HQF-Net. We evaluate HQF-Net by comparing it to several baseline models: state-of-the-art classical models and other hybrid quantum models. 

\subsubsection{Quantitative Results}

We evaluate HQF-Net using two standard semantic segmentation metrics, mean Intersection over Union (mIoU) and Overall Accuracy (OA). Comparative results on LandCover.ai, OpenEarthMap, and SeasoNet are summarized in Tables~\ref{tab:results}, \ref{tab:oem_comparison}, and \ref{tab:seaonet_comparison}.

Table~\ref{tab:results} compares HQF-Net with several classical CNN-based and quantum-inspired architectures on the LandCover.ai dataset. Earlier hybrid quantum models such as FQCNN and MQCNN achieve relatively low performance with mIoU values of 0.2000 and 0.1520 respectively, indicating the difficulty of effectively integrating quantum operations with classical CNNs. Traditional segmentation models including UNet, UNet++, and UNet-SPP significantly improve the results, reaching mIoU scores between 0.64 and 0.69. More advanced architectures, such as DIResUNet, further improve the performance to 0.7522 mIoU and 87.05\% OA. In comparison, the proposed HQF-Net achieves the best performance with an mIoU of \textbf{0.8568} and an OA of \textbf{96.87\%}, demonstrating a consistent improvement over existing classical and hybrid quantum baselines.


\begin{table}[t!]
\centering
\caption{Performance comparison on the LandCover.ai test set.}
\label{tab:results}
\scriptsize
\setlength{\tabcolsep}{4pt}
\begin{tabular}{p{3.1cm} p{2.0cm} c c}
\toprule
\textbf{Source} & \textbf{Model} & \textbf{mIoU} & \textbf{OA (\%)} \\
\midrule
Fan et al.~\cite{fan2024land}            & FQCNN     & 0.2000 & 53.26 \\
Fan et al.~\cite{fan2024land}            & MQCNN     & 0.1520 & 39.03 \\
Kumar et al.~\cite{kumar2024remote}      & CNN       & 0.1500 & 45.87 \\
Kumar et al.~\cite{kumar2024remote}      & COQCNN    & 0.1280 & 36.65 \\
Ronneberger et al.~\cite{ronneberger2015u} & UNet    & 0.6451 & 82.43 \\
Zhou et al.~\cite{zhou2018unet++}        & UNet++    & 0.6553 & 70.89 \\
Abdani et al.~\cite{abdani2020u}         & UNet SPP  & 0.6920 & 71.27 \\
Priyanka et al.~\cite{priyanka2022diresunet} & DIResUNet & 0.7522 & 87.05 \\
\midrule
\textbf{Ours} & \textbf{HQF-Net} & \textbf{0.8568} & \textbf{96.87} \\
\bottomrule
\end{tabular}
\end{table}

\begin{table}[h!]
\centering
\caption{Comparison on OpenEarthMap with State-of-the-Art Methods (mIoU \%)}
\label{tab:oem_comparison}
\setlength{\tabcolsep}{4pt}
\renewcommand{\arraystretch}{0.95}
\begin{tabular}{l l c}
\hline
\textbf{Model} & \textbf{Backbone} & \textbf{mIoU} \\
\hline
UNet \cite{ronneberger2015u}                & ---           & 60.4 \\
DANet \cite{fu2019dual}               & ResNet101     & 60.1 \\
ClipSeg \cite{luddecke2022image}            & CLIP          & 58.6 \\
BoTNet \cite{srinivas2021bottleneck}             & ResNet50      & 61.5 \\
MANet \cite{li2021multiattention}              & ResNet101     & 64.0 \\
SegFormer \cite{xie2021segformer}          & MiT           & 66.0 \\
DC-Swin \cite{wang2022novel}            & Swin-S        & 67.2 \\
UNetFormer \cite{wang2022unetformer}         & Swin-B        & 68.0 \\
ConvNeXt \cite{liu2022convnet}           & ConvNeXt-B    & 64.9 \\
RS$^3$ Mamba \cite{ma2024rs}          & --            & 64.5 \\
PyramidMamba \cite{wang2025pyramidmamba} & ResNet18      & 61.7 \\
                    & Swin-B        & 70.8 \\                   
\midrule
\textbf{HQF-Net (Ours)}             & \textbf{DINOv3-ViT-L16} & \textbf{71.82} \\
\hline
\end{tabular}
\end{table}

\begin{table}[h!]
\centering
\caption{Performance Comparison on SeasoNet (Accuracy and mIoU \%)}
\label{tab:seaonet_comparison}
\setlength{\tabcolsep}{4pt}
\renewcommand{\arraystretch}{0.95}
\begin{tabular}{l l c c}
\hline
\textbf{Model} & \textbf{Backbone} & \textbf{Acc.} & \textbf{mIoU} \\
\hline
DeepLabv3 \cite{kossmann2022seasonet}        & DenseNet121      & --    & 47.53 \\
DeepLabv3 PT \cite{kossmann2022seasonet}     & DenseNet121      & --    & 48.69 \\
DeepLabv3 \cite{chen2017rethinking}        & ResNet-50        & 83.52 & 50.79 \\
DeepLabv3 \cite{chen2017rethinking}        & ConvNeXt-Small   & 81.36 & 46.39 \\
DeepLabv3 \cite{chen2017rethinking}        & Swin-Tiny        & 82.75 & 50.81 \\
UperNet \cite{xiao2018unified}           & ResNet-50        & 83.20 & 49.59 \\
SegFormer \cite{xie2021segformer}        & MiT              & 83.75 & 53.87 \\
\midrule
\textbf{HQF-Net (Ours)}          & \textbf{DINOv3-ViT-L16}     & \textbf{99.37} & \textbf{55.28} \\
\hline
\end{tabular}
\end{table}

The results on the OpenEarthMap benchmark are shown in Table~\ref{tab:oem_comparison}. Several strong baselines including SegFormer, UNetFormer, and ConvNeXt achieve competitive performance with mIoU scores ranging from 64.0\% to 68.0\%. PyramidMamba with a Swin-B backbone achieves 70.8\% mIoU, ranking among the strongest previously reported results. The proposed HQF-Net, equipped with the DINOv3-ViT-L16 backbone, achieves the highest performance of 71.82\% on this benchmark, demonstrating its ability to effectively capture complex spatial patterns in large-scale remote sensing imagery.

\begin{figure*}[t!]
    \centering
    \includegraphics[width=2.1\columnwidth]{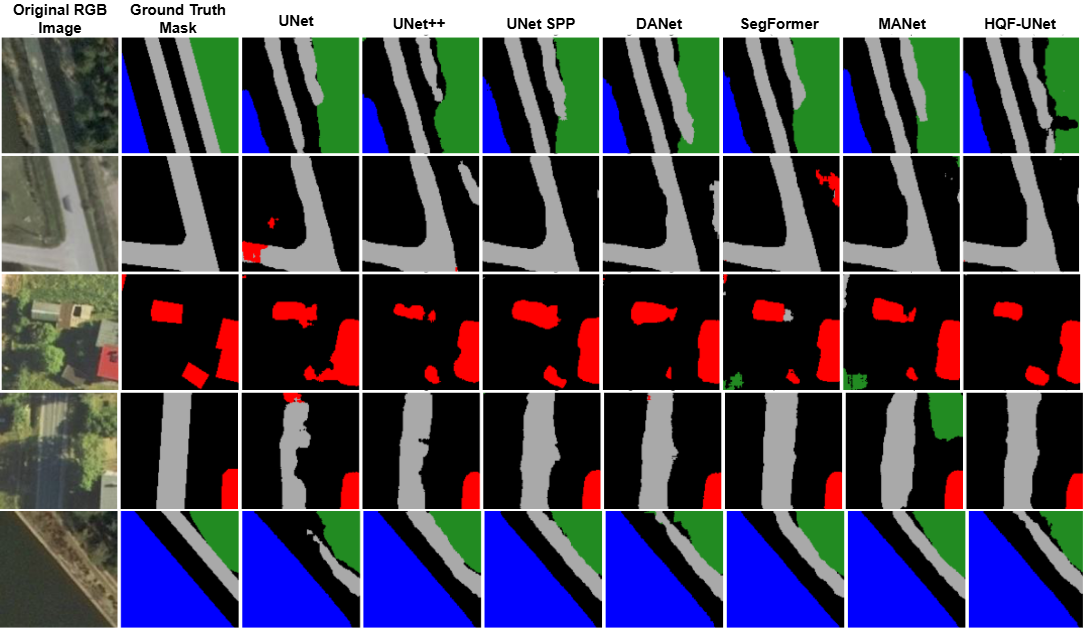}
    \caption{Qualitative segmentation on the LandCover.ai dataset, showing input images, ground-truth masks, and predictions by HQF-Net.}
    \label{fig:lcai_qr}
\end{figure*}

Finally, Table~\ref{tab:seaonet_comparison} reports results on the SeasoNet dataset, which contains challenging multi-temporal agricultural scenes. Existing models such as Deeplabv3 and UperNet achieve mIoU values around 47–50\%, while SegFormer reaches 53.87\% mIoU. HQF-Net slightly improves the mIoU to \textbf{55.28\%} while achieving a significantly higher overall accuracy of \textbf{99.37\%}. The very high OA is partly influenced by class imbalance in the dataset. These results indicate that the proposed hybrid quantum feature processing modules effectively enhance feature representation and improve segmentation performance across diverse remote sensing datasets.

HQF-Net achieves strong and consistent performance across three remote sensing semantic segmentation benchmarks. It achieved \textbf{0.8568} mIoU and \textbf{96.87\%} OA on LandCover.ai, and \textbf{71.82\%} mIoU on OpenEarthMap using a \textbf{DINOv3-ViT-L16} backbone. On the more challenging multi-temporal SeasoNet benchmark, HQF-Net achieves \textbf{55.28\%} mIoU and \textbf{99.37\%} accuracy, showing a modest but consistent gain over the strongest baseline. Overall, these results suggest that HQF-Net generalizes effectively across diverse remote sensing scenarios and benefits from the integration of multi-scale semantic fusion and quantum-enhanced feature processing.

\subsubsection{Qualitative Analysis}

\begin{table*}[t]
\centering
\caption{Ablation study showing the impact of different fusion strategies and quantum components on the LandCover.ai dataset.}
\label{tab:ablation_arch}
\setlength{\tabcolsep}{5pt} 
\renewcommand{\arraystretch}{1.1} 
\begin{tabular}{lcccccc}
\toprule
\textbf{Model Variant} & \textbf{DINOv3} & \textbf{DMCAF} & \textbf{Q-Skip} & \textbf{Q-MoE} & \textbf{mIoU} & \textbf{OA (\%)} \\
\midrule
\multicolumn{7}{l}{\textbf{\textit{Baseline Fusion Methods}}} \\
U-Net + DINOv3 (Multiplication)   & \checkmark &            &            &            & 0.7335 & 90.24 \\
U-Net + DINOv3 (Addition)         & \checkmark &            &            &            & 0.7520 & 91.80 \\
\midrule
\multicolumn{7}{l}{\textbf{\textit{Proposed Hybrid Architectures}}} \\
+ DMCAF (Deformable Fusion)       & \checkmark & \checkmark &            &            & 0.7815 & 92.10 \\
+ DMCAF + Q-Skip                  & \checkmark & \checkmark & \checkmark &            & 0.8387 & 94.22 \\
+ DMCAF + Q-MoE                   & \checkmark & \checkmark &            & \checkmark & 0.8429 & 94.78 \\
\textbf{HQF-Net (Full Model)}     & \checkmark & \checkmark & \checkmark & \checkmark & \textbf{0.8568} & \textbf{96.87} \\
\bottomrule
\end{tabular}
\end{table*}
This section presents qualitative comparisons of segmentation results between HQF-Net and several baseline models on representative samples from the LandCover.ai dataset. Additional qualitative results for OpenEarthMap and SeasoNet are provided in the supplementary material. The results highlight the proposed architecture's ability to produce more accurate, spatially consistent segmentation masks.

On the LandCover.ai dataset, as shown in Fig.~\ref{fig:lcai_qr}, traditional architectures such as UNet, UNet++, and UNet-SPP capture the general structure of large objects but often struggle with fine boundaries and small regions. In contrast, HQF-Net produces cleaner object boundaries and preserves smaller structures such as narrow roads and building edges. Compared with attention-based models such as DANet \cite{fu2019dual} and MANet \cite{li2021multiattention}, the proposed method reduces misclassified regions and produces more coherent segmentation maps.

Additional qualitative examples for OpenEarthMap and SeasoNet are provided in the supplementary material. Overall, the qualitative results show that HQF-Net produces cleaner boundaries, more consistent segmentation maps, and better recognition of both small and large objects across different remote sensing datasets, even though a few minor artefacts remain, such as slight confusion between the Woodland and Water classes and small thickening of road segments at complex intersections. These visual improvements are consistent with the quantitative gains and indicate that the proposed fusion and quantum refinement modules help maintain both semantic consistency and spatial structure.

\subsubsection{Ablation Study}
To quantify the contributions of the main components in HQF-Net, we conduct an architectural ablation study on the LandCover.ai dataset using the same training protocol as for the full model. Starting from naive fusion strategy on a classical U-Net baseline, we progressively incorporate multi-scale DINOv3 feature guidance, DMCAF, QSkip, and QMoE, as shown in Table~\ref{tab:ablation_arch}.


Overall, the ablation results validate the design of HQF-Net. Each component contributes incrementally, with DMCAF providing the largest gain, while QSkip and QMoE further refine performance. These findings show that the complementary integration of DINOv3 guidance, DMCAF-based feature fusion, quantum-enhanced skip refinement, and the QMoE bottleneck is effective for RS semantic segmentation.
\section{Conclusion}
\label{sec:conc_fw}

We introduced HQF-Net, a hybrid quantum--classical multi-scale fusion network for remote sensing semantic segmentation. HQF-Net combines DINOv3-guided semantic fusion, DMCAF-based feature alignment, quantum-enhanced skip refinement, and a QMoE bottleneck within a unified U-Net-style architecture. This design enables the model to capture complementary local, global, and directional feature dependencies for dense prediction. Across three remote sensing segmentation benchmarks, HQF-Net achieves strong and consistent performance, including \textbf{0.8568} mIoU and \textbf{96.87\%} OA on LandCover.ai, \textbf{71.82\%} mIoU on OpenEarthMap, and \textbf{55.28\%} mIoU with \textbf{99.37\%} overall accuracy on SeasoNet. The ablation study further validates the contribution of each major component in the final architecture. Overall, the results suggest that structured hybrid quantum-classical feature processing is a promising direction for remote sensing image segmentation under near-term quantum constraints. As future work, we plan to extend HQF-Net to larger remote sensing benchmarks and investigate quantum-based self-supervised representation learning.


{
  \small
  \bibliographystyle{ieeenat_fullname}
  \bibliography{references}
}

\clearpage
\appendix
\setcounter{page}{1}
\maketitlesupplementary

\section{Quantum Operations Used in HQF-Net}
\label{sec:qc_background_s}

HQF-Net employs parameterized quantum circuits to transform compact classical feature representations in the bottleneck and skip pathways. A qubit state can be represented as a superposition of basis states, allowing compressed feature responses to be encoded in a higher-dimensional Hilbert space. Parameterized rotation gates, such as $R_Y(\theta)$, are used to introduce learnable transformations, while CNOT-based entanglement operations model interactions between qubits. In HQF-Net, these quantum operations are not used as a standalone quantum model; instead, they function as feature enrichment modules embedded within a hybrid encoder--decoder segmentation pipeline. This design enables the network to capture structured correlations in compressed latent features while preserving compatibility with classical convolutional processing.

To clarify the role of these quantum modules in HQF-Net, we briefly review the quantum concepts that underpin their design. Specifically, qubit state representations provide the basis for encoding compressed classical features, parameterized quantum gates define learnable transformations over these encoded states, and entanglement enables the modeling of correlations between latent feature components. These concepts form the foundation of the quantum operations used in both the QMoE bottleneck and QSkip pathways.

\subsection{Qubit \& Hilbert Space Representation}
The fundamental unit of quantum information is the qubit, analogous to a classical bit but with the unique property of being able to exist in a superposition \cite{hughes2020introduction,grover1998advantages} of states. A qubit's state \( |\psi\rangle \) is represented as a linear combination of the basis states \( |0\rangle \) and \( |1\rangle \):
\[
|\psi\rangle = \alpha |0\rangle + \beta |1\rangle
\]
where \( \alpha \) and \( \beta \) are complex probability amplitudes satisfying the normalization condition \( |\alpha|^2 + |\beta|^2 = 1 \). The ability of qubits to exist in superposition enables quantum systems to encode and process information in exponentially larger representational spaces than classical systems.

A system of \( Q \) qubits is described in a tensor product Hilbert space, \( \mathcal{H}_{2^Q} = \mathcal{H}_2^{\otimes Q} \), whose dimensionality grows exponentially with \( Q \). For instance, a two-qubit system can exist in a superposition of the four states \( |00\rangle, |01\rangle, |10\rangle, |11\rangle \), providing a powerful basis for quantum computation.

\subsection{Quantum Gates as Unitary Rotations}
Quantum gates are the operations that manipulate qubits. These gates perform unitary transformations that evolve a qubit's state, and they can be visualized as rotations on the Bloch sphere \cite{glendinning2005bloch}. The primary gates used in our approach are:

\begin{enumerate}
    \item \textbf{Hadamard Gate (H)}: This gate \cite{shepherd2006role} creates superposition by transforming a basis state into an equal superposition of \( |0\rangle \) and \( |1\rangle \).
    \[
    H = \frac{1}{\sqrt{2}}
    \begin{pmatrix}
    1 & 1 \\
    1 & -1
    \end{pmatrix}; \quad H|0\rangle = \frac{1}{\sqrt{2}}(|0\rangle + |1\rangle)
    \]
    
    \item \textbf{Rotation Gates}: These parameterized gates \cite{du2020expressive} rotate the state vector around an axis of the Bloch sphere by an angle \( \theta \). These gates are used to introduce learnable transformations and enable flexible feature encoding.
    \[
    R_Y(\theta) = \begin{pmatrix}
    \cos(\theta/2) & -\sin(\theta/2) \\
    \sin(\theta/2) & \cos(\theta/2)
    \end{pmatrix}
    \]
\end{enumerate}

The use of these gates \cite{williams2011quantum} allows quantum models to perform transformations on feature representations in ways that may capture intricate interactions that are difficult to model efficiently with standard classical operations.

\subsection{Entanglement and Multi-Qubit Gates}
Entanglement \cite{horodecki2009quantum} is a unique quantum phenomenon that describes correlations between qubits that cannot be separated into individual qubit states. These correlations enable quantum circuits to model complex feature interactions in a different representational space from classical systems. The \textbf{Controlled-NOT (CNOT)} gate \cite{wong2023quantum} is commonly used to create entanglement between qubits:
\[
\text{CNOT}|10\rangle = |11\rangle; \quad \text{CNOT}|01\rangle = |01\rangle
\]
By applying a Hadamard gate followed by a CNOT gate, we can create the canonical entangled Bell state:
\[
|\Phi^+\rangle = \frac{1}{\sqrt{2}} (|00\rangle + |11\rangle)
\]
In our model, entanglement is used within quantum circuits at key points in the U-Net architecture, such as the bottleneck, to enhance feature extraction and representation learning. These quantum circuits enable the model to capture high-level correlations across multi-resolution features, which can be beneficial for segmentation tasks.

\subsection{Quantum Approaches in Deep Learning}
Quantum models have recently been integrated into classical deep learning architectures, where quantum circuits are placed at strategic locations \cite{sebastianelli2021circuit, otgonbaatar2021classification}, such as the bottleneck layer \cite{de2024towards, islam2025qpolypnet}, to enhance feature extraction. The quantum bottleneck approach has shown promise by incorporating quantum circuits at the highest level of feature compression, thereby facilitating better representation learning. Li et al. \cite{li2022image} applied a hybrid model combining classical feature extractors with quantum classifiers, demonstrating that quantum circuits can complement classical architectures in image classification tasks.

By combining classical architectures with quantum circuits \cite{miroszewski2023quantum}, hybrid quantum-classical models leverage the strengths of both: the flexibility of classical deep learning for general tasks and the expressiveness of quantum circuits for complex feature interactions. Building on this hybrid paradigm, HQF-Net inserts quantum modules at two complementary locations: the skip pathways for intermediate feature refinement and the bottleneck for compact latent transformation through expert-based routing.
 
\section{Quantum Feature Processing Modules}
In HQF-Net, classical feature maps are first projected into a compact latent representation and encoded into quantum states. Structured parameterized circuits are then used to transform these states, and the resulting measurements are projected back to classical feature space. These quantum modules are used in two parts of the architecture: the QSkip pathways and the QMoE bottleneck. The QSkip branch uses a multi-scale enrichment circuit to refine skip features, while the QMoE module routes latent features through specialized quantum experts designed to capture complementary local, global, and directional dependencies.

1. \textbf{Enrichment Multi-Scale Circuit.}
This circuit is designed to enhance image features by incorporating both global and local feature correlations. The circuit consists of local convolutional layers that apply localized operations on neighboring qubits to extract fine-grained spatial patterns. Then a global mixing layer that entangles all qubits propagates information across the entirety of the input. This design enables the circuit to preserve very detailed, structured information locally while maintaining the overall structure globally, thereby allowing it to capture and represent many different types of patterns in the data. In this architecture, the circuit is utilized in our QSkip module and before the QMoE block.

2. \textbf{Localist Expert.}
 The Localist Circuit captures local and fine-grained features. This circuit is a simple quantum circuit that has entanglements only between neighboring qubits. In effect, this localist circuit mimics a classical two-dimensional separable convolution operator and allows the capture of small spatial correlations (e.g., an edge, corner, localized texture). Thus its principal benefit is the retrieval of fine detail that would otherwise not be retrieved from any of the deeper encoder layers or from global operations.

3. \textbf{Globalist Expert.}
The Globalist circuit captures long-distance dependencies as well as the entire range of global patterns across an entire feature map. The Globalist circuit employs no local convolutions, but rather utilizes deep entanglement of qubits and rotation gates. It also creates the ability to model correlations between non-adjacent qubits, allowing the circuit to capture repeating patterns; assigns symmetry to data; and provides a model for capturing long-term structural dependencies. The Globalist circuit complements the Localist expert by encoding relevant global context that cannot be encoded solely through local operations or local connectivity.

4. \textbf{Diagonal Expert.}
The Diagonal circuit captures structured, directional dependencies that cannot be represented as either a purely local or a purely global circuit. By using parameterized rotation gates followed by CNOT gates on non-adjacent diagonally indexed qubits, this circuit uses a staggered entanglement topology, allowing it to model diagonal or directional relationships among the features in the image, thus creating a structure to help the system learn to identify complex patterns in the features of the image based on directionality and structure.

5. \textbf{Two-Qubit Quantum Filter.}
The convolution-like operation is implemented using a two-qubit parameterized unitary, as illustrated in Figure~1. This unit serves as a basic local interaction block within the larger circuit designs used in HQF-Net.

\begin{figure}[h!]
    \centering
    \includegraphics[width=0.3\linewidth]{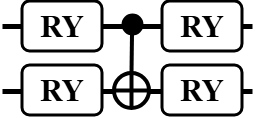}
    \caption{Two-qubit parameterized quantum filter used as a local interaction block in HQF-Net. The unit models pairwise feature interactions and serves as a basic building block within the larger quantum circuit designs.}
    \label{fig:conv_filter}
\end{figure}

\section{Data Pre-processing and Pipeline}
For LandCover.ai and OpenEarthMap, we use random cropping, which samples $224\times224$ patches from the larger source tiles. For SeasoNet, which has a native patch size of 120 × 120, all input images/masks were resized using bilinear/nearest-neighbour interpolation to $224\times224$ so that they have the same spatial dimensions and can therefore be processed through the encoder and decoder blocks. Finally, each patch is normalised to the range $[0,1]$.

\section{Model Architectural Discussion}
HQF-Net combines semantic-guided fusion, multi-scale quantum enrichment, and expert-based quantum routing within a U-Net-style encoder-decoder framework. DMCAF injects semantically informed context into encoder features while preserving spatial fidelity through sparse deformable sampling. The bottleneck compresses the high-dimensional encoder output into a compact latent representation, which is enriched through structured quantum transformations before adaptive routing across specialized experts. In parallel, quantum-enhanced skip pathways refine intermediate features passed to the decoder. Together, these components allow HQF-Net to combine local detail preservation, global context modeling, and adaptive feature specialization for remote sensing segmentation.

\section{Qualitative Results for OpenEarthMap and SeasoNet}
For OpenEarthMap scenes shown in Fig.~\ref{fig:oem_model_qr}, which contain complex urban layouts and heterogeneous land-cover patterns, HQF-Net demonstrates improved delineation of buildings, roads, and surrounding land areas. Baseline methods sometimes produce fragmented predictions or confuse neighboring classes, particularly in dense urban regions. HQF-Net generates smoother and more complete segmentation maps while maintaining sharp structural boundaries. From the visual comparisons, conventional architectures such as U-Net tend to produce coarser segmentation masks and occasionally miss small structures or thin road segments. Transformer-based models such as SegFormer, DC-Swin, and UNetFormer improve the overall structural representation; however, they still produce fragmented predictions or minor misclassifications around object boundaries and densely packed regions.

\begin{figure*}[h!]
    \centering
    \includegraphics[width=2.1\columnwidth]{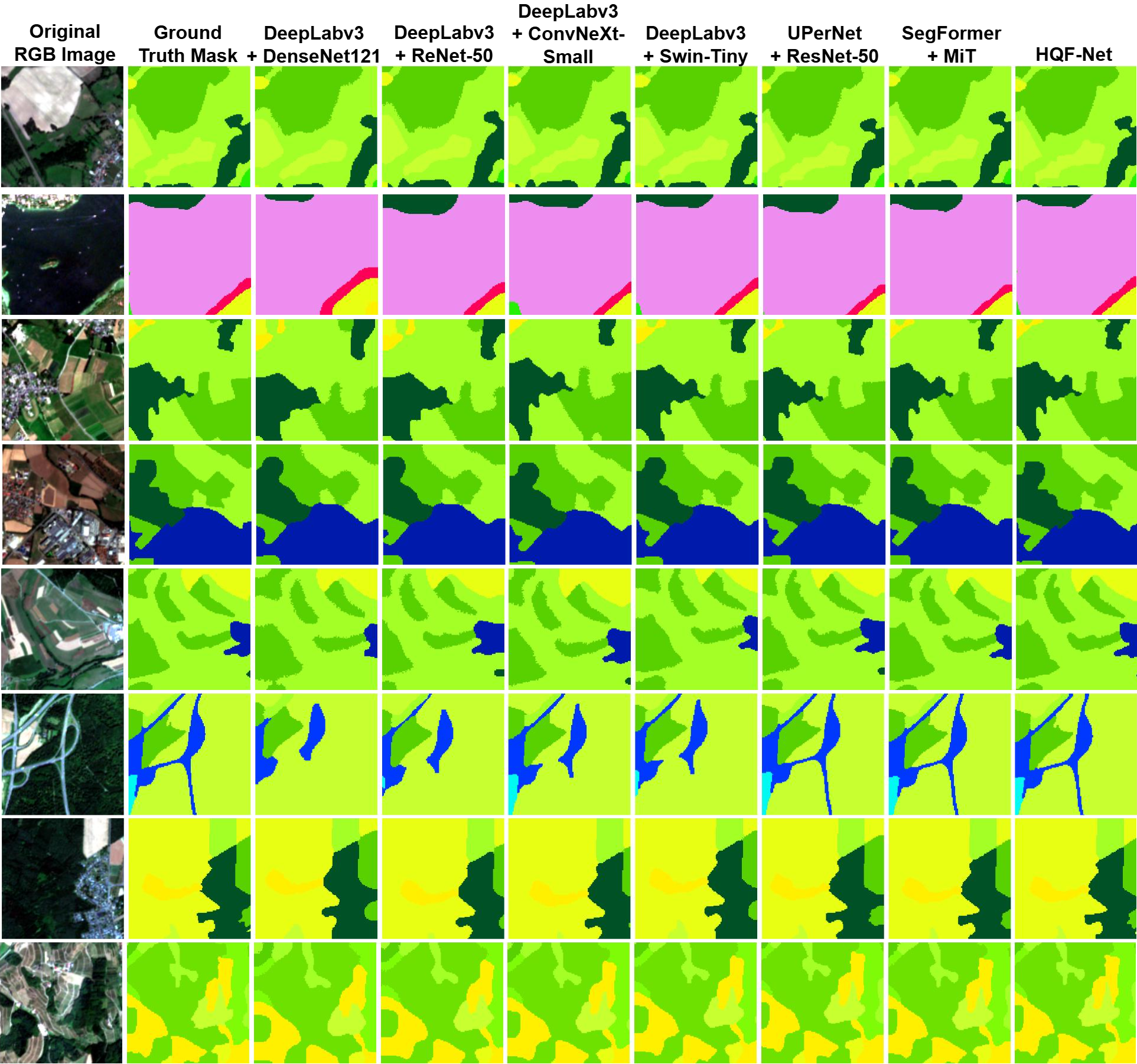}
    \caption{Qualitative segmentation results on the SeasoNet dataset showing original images, ground-truth masks, and comparisons with other models.}
    \label{fig:sn_model_qr}
\end{figure*}

\begin{figure*}[h!]
    \centering
    \includegraphics[width=2.1\columnwidth]{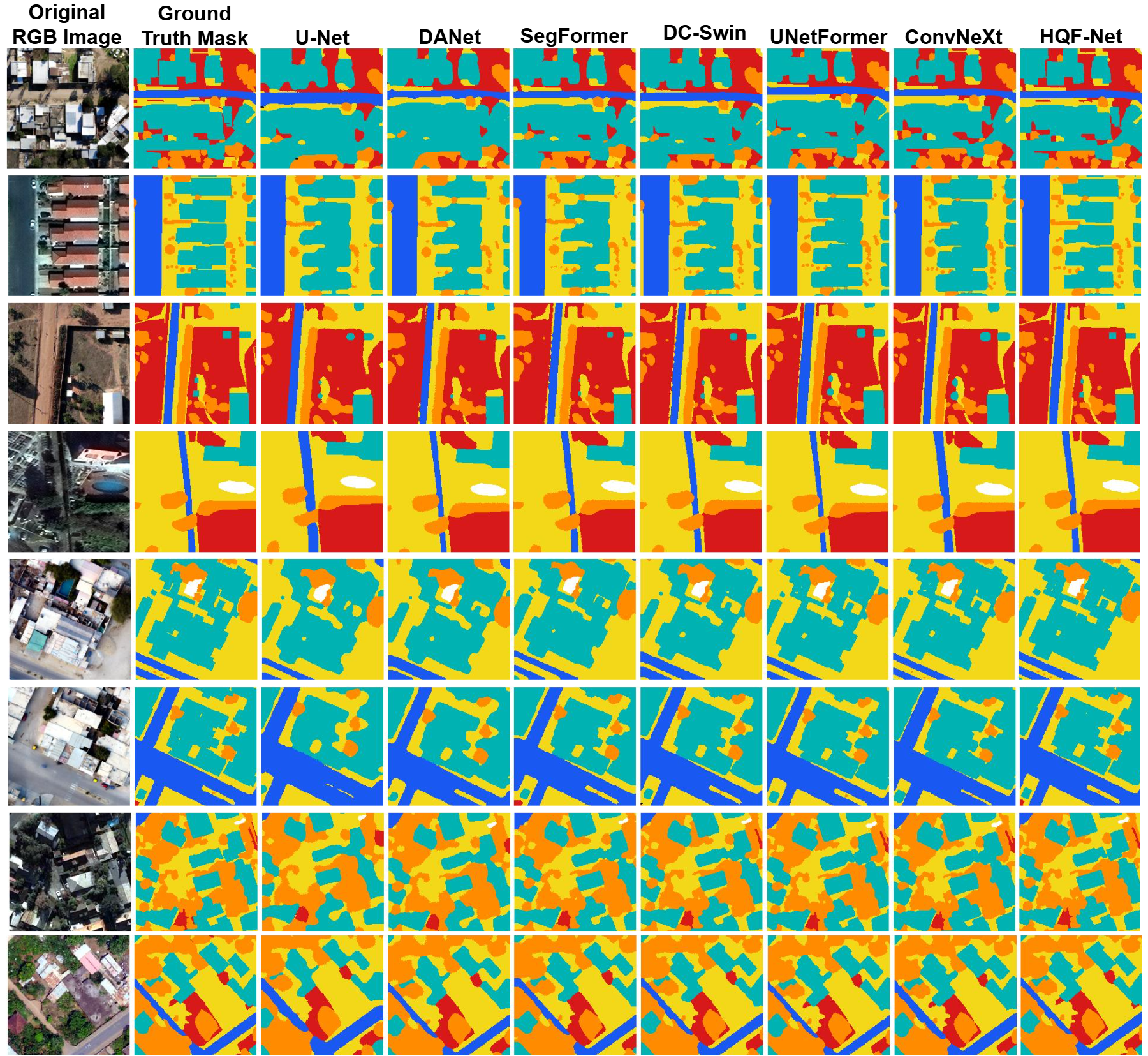}
    \caption{Qualitative segmentation results on the OpenEarthMap dataset showing original images, ground-truth masks, and comparisons with other models.}
    \label{fig:oem_model_qr}
\end{figure*}

Figure~\ref{fig:sn_model_qr} presents qualitative comparisons between HQF-Net and several baseline models on the SeasoNet dataset. This dataset contains multi-temporal agricultural scenes with complex vegetation patterns and subtle class transitions, making accurate segmentation particularly challenging. From the visual results, DeepLabv3-based models with different backbones (DenseNet121, ResNet-50, ConvNeXt-Small, and Swin-Tiny) generally capture the large land-cover regions but often produce coarse boundaries and slight inconsistencies between adjacent classes. Similarly, UPerNet and SegFormer provide improved spatial representation but occasionally introduce minor boundary artifacts or fragmented regions, especially around irregular vegetation patterns and water bodies.

\begin{figure*}[h!]
    \centering
    \includegraphics[width=2.1\columnwidth]{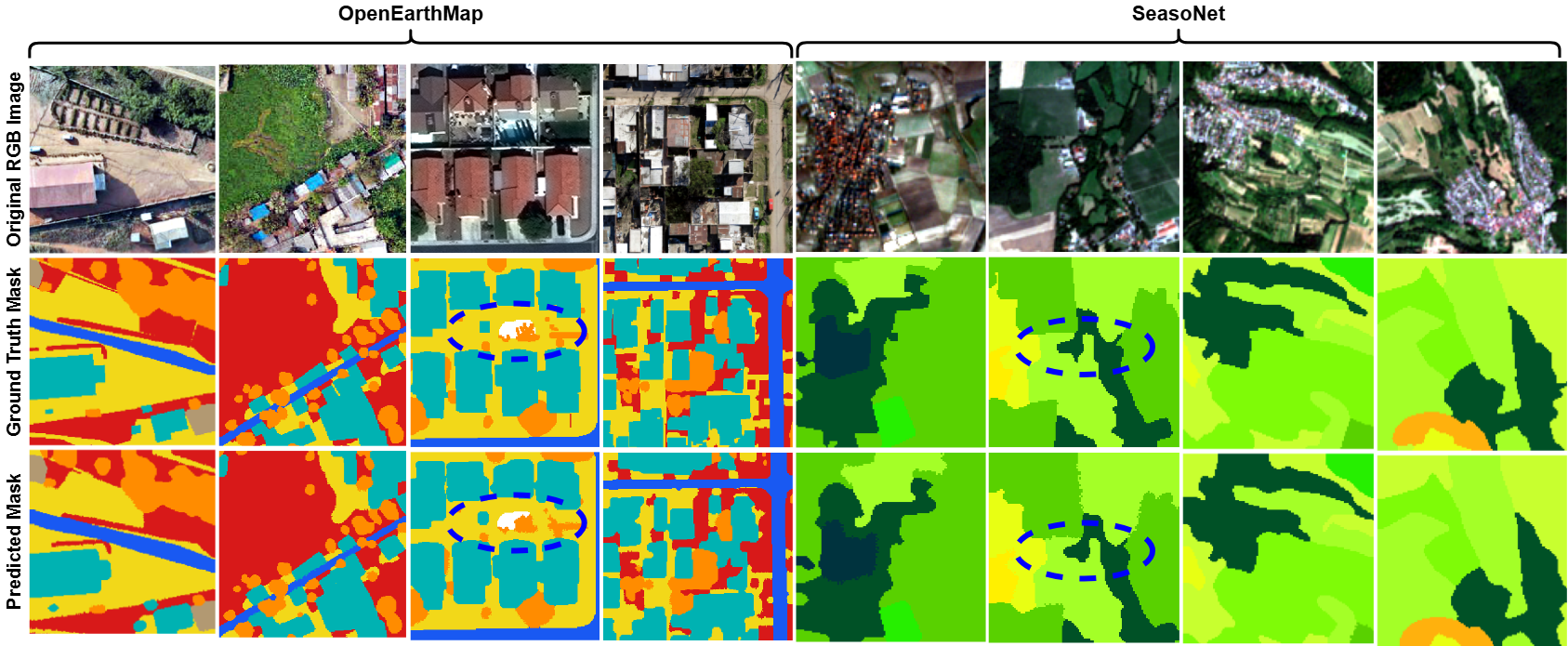}
    \caption{Qualitative segmentation results on the OpenEarthMap and SeasoNet datasets showing original images, ground-truth masks, and HQF-Net predictions.}
    \label{fig:oem_sn_qr}
\end{figure*}

Figure~\ref{fig:oem_sn_qr} shows representative qualitative segmentation results of HQF-Net on the OpenEarthMap and SeasoNet datasets. The examples demonstrate challenging urban and semi-urban scenes with complex spatial layouts, including dense buildings, roads, and mixed land cover.

Overall, the qualitative results indicate that HQF-Net provides more stable and spatially coherent predictions on the SeasoNet dataset, effectively capturing both large homogeneous regions and finer structural details in complex agricultural environments.

\section{Quantum Circuit Implementation and Simulation Environment}
\label{ssec:implementation_details}

All quantum components of our architecture were designed and simulated using \textbf{PennyLane} \cite{bergholm2018pennylane}, a widely used quantum differentiable programming library. PennyLane's seamless integration with \textbf{PyTorch} was instrumental, allowing the quantum circuits to be inserted directly into the classical deep learning model as trainable layers. This enabled fully end-to-end training of the entire hybrid system using standard backpropagation.

All quantum simulations were run on classical hardware using PennyLane's high-performance backends in order to ensure computational feasibility and to speed up our experiments. We used the \texttt{lightning.gpu} device for GPU execution. These backends provide much greater speed when compared to pure Python. Furthermore, in order to speed up the training process, we used the adjoint differentiation method to compute the gradients of the trainable parameters in the quantum circuits. Compared to the standard parameter-shift rule, adjoint differentiation is substantially quicker for simulation, as it computes all of the gradients with a constant, small number of circuit executions, allowing backpropagation through the quantum layers to be very efficient.

\section{Limitations and Future Directions}
\label{sec:limitations}

While our proposed hybrid quantum architectures show competitive performance, this study has several limitations that highlight key challenges and pave the way for future research in hybrid quantum-classical computer vision.

\subsection{Computational Cost of Quantum Simulation}

The most significant practical limitation of this work is the computational cost associated with simulating quantum circuits. Although our final architectures were designed to reduce the number of quantum circuit evaluations per forward pass, the training time was still substantially longer than that of a purely classical model. The overhead of initializing the quantum simulator, constructing the circuit, and performing state vector simulation, even on a high-performance backend like \texttt{lightning.gpu}, remains a major bottleneck. This underscores the critical need for either more optimized classical simulators or, ultimately, fault-tolerant quantum hardware to make the training of such deep hybrid models truly scalable.

\subsection{Challenges in Hybrid Model Training and Debugging}
Developing and training our hybrid models that are more fully integrated posed challenges beyond those of standard deep learning pipelines. All of these models were able to converge; however, maintaining stable and consistent gradient flow through the quantum layers was non-trivial. This required careful initialization and the use of adjoint differentiation to maintain stable training. However, as quantum circuits become more sophisticated and deeper, it is likely that they will also face greater optimization difficulties, such as barren plateaus, than what has been studied here.

\subsection{Simulation vs.\ Real-World Hardware}
All experiments in this study were conducted on classical computers using quantum simulation. As a result, the findings are approximate representations of how quantum modules would behave under optimal conditions, rather than real quantum hardware. Assessing the performance of these designs on current NISQ platforms remains an unresolved problem, due to gate errors, qubit decoherence, measurement noise, and limited qubit interconnectivity in today's physical implementations. Additional research on error mitigation, hardware-aware circuit transpilation, and noise-robust quantum module design would be needed for practical empirical deployment. Therefore, the reported results should be interpreted as an optimistic estimate of the potential behavior of these architectures within the simulated backends.

\subsection{Limited Architectural Search}
In summary, although we explored several novel architectural designs, the design space for hybrid quantum-classical architectures remains very large. The quantum expert circuits, DINOv3 fusion technique, and U-Net base are merely one point in a very wide design space. More research that examines a larger area will generate other quantum circuit designs, different attention-based algorithms, and various forms of classical backbones resulting in better performance than we have demonstrated in this research paper. This work introduces one promising architecture and evaluates its capabilities across several settings, but it does not claim to exhaust the broader design space.

\end{document}